\documentclass[letterpaper]{article} 
\usepackage{aaai23}  
\usepackage{times}  
\usepackage{helvet}  
\usepackage{courier}  
\usepackage[hyphens]{url}  
\usepackage{graphicx} 
\usepackage{natbib}  
\usepackage{caption} 
\usepackage{algorithm}
\usepackage{algorithmic}
\usepackage{times}
\usepackage{helvet}
\usepackage{courier}
\usepackage{microtype}
\usepackage{amsfonts}
\usepackage{amssymb}
\usepackage{multirow}
\usepackage{graphicx}
\usepackage{amssymb}
\usepackage{natbib}
\usepackage{amsmath}
\usepackage{appendix}
\usepackage{subfigure}
\usepackage{booktabs}
\usepackage{algorithm}
\usepackage{algorithmic}

\usepackage{newfloat}
\usepackage{listings}
\DeclareCaptionStyle{ruled}{labelfont=normalfont,labelsep=colon,strut=off} 
\floatstyle{ruled}
\newfloat{listing}{tb}{lst}{}
\floatname{listing}{Listing}
\title{Generalized Category Discovery with Decoupled Prototypical Network}
\author{
    Wenbin An\textsuperscript{\rm 1,3}, Feng Tian\textsuperscript{\rm 2,3}, Qinghua Zheng\textsuperscript{\rm 2,3}, Wei Ding\textsuperscript{\rm 4}, QianYing Wang\textsuperscript{\rm 5}, Ping Chen\textsuperscript{\rm 6}\\
}
\affiliations{
    \textsuperscript{\rm 1}School of Automation Science and Engineering, Xi'an Jiaotong University \\
    \textsuperscript{\rm 2}School of Computer Science and Technology, Xi'an Jiaotong University \\
    \textsuperscript{\rm 3}National Engineering Laboratory for Big Data Analytics \\
    \textsuperscript{\rm 4}Department of Computer Science, University of Massachusetts Boston \\
    \textsuperscript{\rm 5}Lenovo Research \\
    \textsuperscript{\rm 6}Department of Engineering, University of Massachusetts Boston\\
    wenbinan@stu.xjtu.edu.cn, \{fengtian,qhzheng\}@mail.xjtu.edu.cn \\
    \{wei.ding,Ping.Chen\}@umb.edu, wangqya@lenovo.com
}

\usepackage{bibentry}

\begin{document}
\maketitle

\begin{abstract}
Generalized Category Discovery (GCD) aims to recognize both known and novel categories from a set of unlabeled data, based on another dataset labeled with only known categories. 
Without considering differences between known and novel categories, current methods learn about them in a coupled manner, which can hurt model's generalization and discriminative ability. Furthermore, the coupled training approach prevents these models transferring category-specific knowledge explicitly from labeled data to unlabeled data, which can lose high-level semantic information and impair model performance.
To mitigate above limitations, we present a novel model called \textit{Decoupled Prototypical Network} (DPN).
By formulating a bipartite matching problem for category prototypes, DPN can not only decouple known and novel categories to achieve different training targets effectively, but also align known categories in labeled and unlabeled data to transfer category-specific knowledge explicitly and capture high-level semantics.
Furthermore, DPN can learn more discriminative features for both known and novel categories through our proposed \textit{Semantic-aware Prototypical Learning} (SPL).
Besides capturing meaningful semantic information, SPL can also alleviate the noise of hard pseudo labels through semantic-weighted soft assignment.
Extensive experiments show that DPN outperforms state-of-the-art models by a large margin on all evaluation metrics across multiple benchmark datasets. Code and data are available at \url{https://github.com/Lackel/DPN}.
\end{abstract}

\section{Introduction}
Although modern machine learning methods have achieved superior performance on many NLP tasks such as text classification, they usually fail to recognize novel categories from newly collected unlabelled data. 
To cope with this limitation, some task settings were proposed to discover novel categories from unlabeled text \citep{thu2020} or images \citep{selflabel}. 
Recently, \citet{gcd} comprehensively formalized these settings and proposed a task called Generalized Category Discovery (GCD).
Since models trained on a labeled dataset containing only known categories may encounter newly collected unlabelled data with both known and novel categories, GCD requires models to recognize both known and novel categories from the unlabeled data without any additional annotation, which can help to extend existing category taxonomy and reduce significant labeling cost.


To discover novel categories from unlabeled data, current methods \citep{thu2020, thu2021, gcd} usually employ two steps: pretraining on labeled data for representation learning and then pseudo-label training on unlabeled data to discover categories.
Although these methods can learn some discriminative features for novel categories, they usually face following limitations.
First, these methods cannot transfer category-specific knowledge explicitly from labeled data to unlabeled data, which can lose high-level semantic information and impair model performance on known categories. Even though these methods can transfer some general knowledge implicitly through the pretrained feature extractor, they usually fail to transfer category-specific knowledge since they discard the pretrained classifier after pre-training and cannot align known categories in labeled and unlabeled data.
Second, these methods cannot decouple known and novel categories from unlabeled data to achieve different training goals for them, which can hurt model's generalization and discriminative ability. Without considering different training objectives for known and novel categories, these methods simply treat them the same and learn about them in a coupled unsupervised manner, which can prevent models learning about discriminative features for both known and novel categories. 

In this paper, we first provide insight of different training objectives for known and novel categories in GCD, which is ignored by previous methods. Since we have both labeled and unlabeled data for known categories, we want to \textbf{exploit} knowledge acquired from labeled data to improve model's \textbf{generalization ability} on these categories in a \textbf{semi-supervised manner}. Meanwhile we have only unlabeled data for novel categories, we mainly want to \textbf{explore} some novel knowledge to improve model's \textbf{discriminative ability} on these categories in an \textbf{unsupervised manner}. So learning about known and novel categories with different training approaches in a decoupled manner is crucial for GCD.

To cope with above challenges, we propose \textit{Decoupled Prototypical Network} (DPN), a novel model which can transfer category-specific knowledge explicitly and decouple known and novel categories to achieve different training goals for them.
First, we formulate a bipartite matching problem for category prototypes to align known categories in labeled and unlabeled data, meanwhile decouple known and novel categories from unlabeled data, without introducing any additional model parameters.
Specifically, we first learn a set of category prototypes for labeled data to represent known categories and another set of category prototypes for unlabeled data to represent both known and novel categories. Then we align these two sets of prototypes by formulating it as a bipartite matching problem, which can be solved efficiently with Hungarian algorithm \citep{hungarian}. And the matched prototypes in unlabeled data correspond to known categories and the unmatched prototypes correspond to novel categories.
In this way, DPN can decouple known and novel categories from unlabeled data and learn about them with different training manners to achieve different training objectives.
Furthermore, DPN can utilize the prototypes learned from labeled data to guide the pseudo-label training process on unlabeled data in a category-specific way. So DPN can transfer not only general knowledge through the pre-trained feature extractor, but also category-specific knowledge through prototypes to capture high-level semantics and improve model performance on known categories. 

To learn more discriminative features for both known and novel categories, we further propose \textit{Semantic-aware Prototypical Learning} (SPL). Different from traditional prototypical learning \citep{proto,pcl} which assigns each instance into a single prototype based on noisy pseudo labels, SPL assigns instances into prototypes in a soft manner based on semantic similarities between them, which can acquire meaningful semantic knowledge and alleviate the effect of pseudo-label noise.
Finally, we update the prototypes learned from labeled data through \textit{Exponential Moving Average} (EMA) algorithm. In addition to utilizing knowledge obtained from labeled data through prototypes, we also iteratively update these prototypes to capture the latest acquired knowledge so that it can generalize better to unseen instances. And the EMA algorithm can yield more consistent representations for prototypes to make the training process more stable.

Our main contributions can be summarized as follows:
\begin{itemize}
  \item We propose a novel \textit{Decoupled Prototypical Network} (DPN) for Generalized Category Discovery. By formulating a bipartite matching problem for category prototypes, DPN can decouple known and novel categories to achieve different training goals for them and transfer category-specific knowledge to capture high-level semantics.
  \item We propose \textit{Semantic-aware Prototypical Learning} (SPL) to learn more discriminative features, which can capture semantic relationships between instances and prototypes while alleviating the effect of pseudo-label noise.
  \item Extensive experiments on multiple benchmark datasets show that our model outperforms state-of-the-art methods by a large margin.
\end{itemize}
\begin{figure*}
\centering
\includegraphics[width=17cm, height=6.4cm]{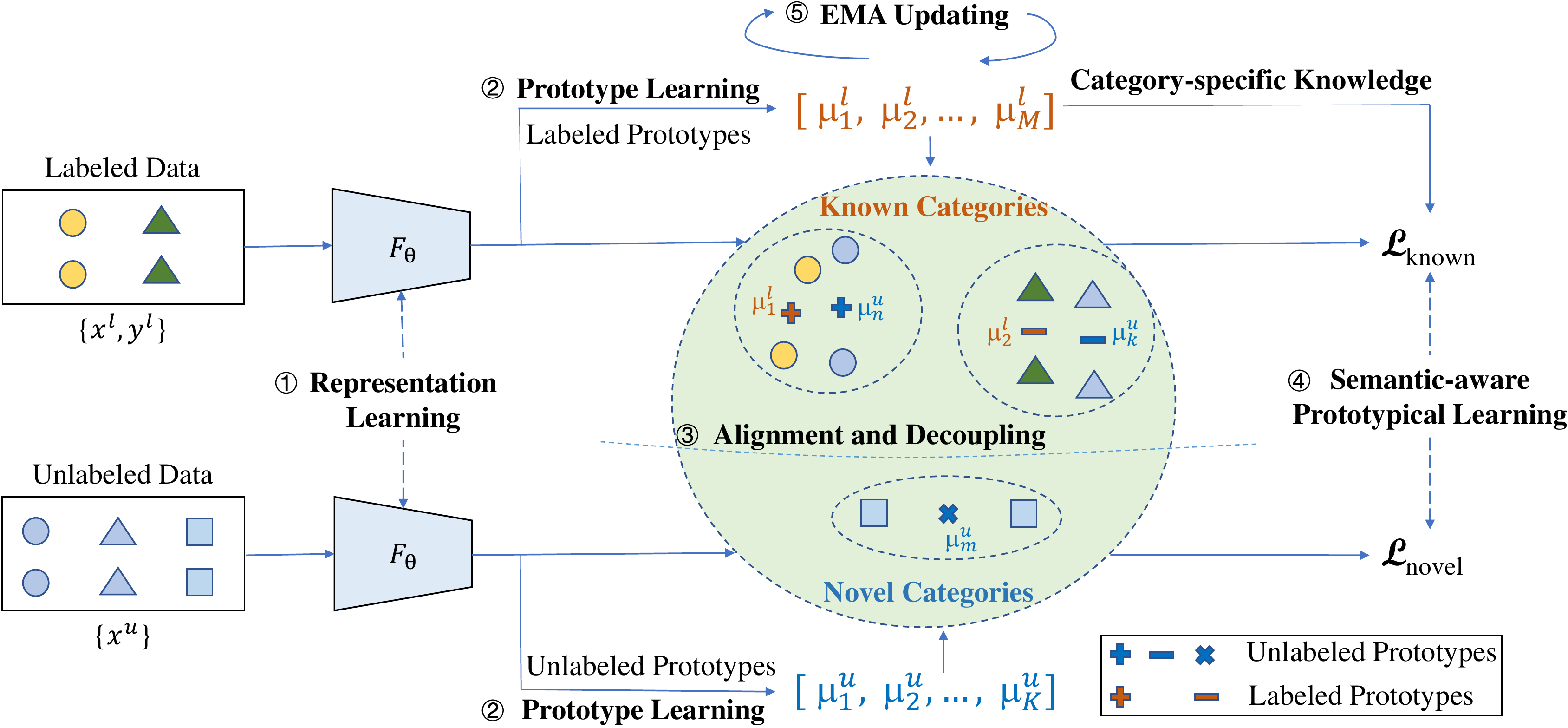}
\caption{An overview of our model.} 
\label{fig:picture1}
\end{figure*}

\section{Related Work}
\subsection{Generalized Category Discovery}
Generalized Category Discovery (GCD) was first comprehensively formalized by \citet{gcd}. They further proposed a simple baseline to use contrastive learning and semi-supervised learning to discover novel categories. Another line of work aimed to discover novel user intents from unlabeled utterances, and they usually adopted pseudo-labeling methods to train their models. For example, \citet{thu2020} adopted pseudo supervision based on pairwise similarities to guide the clustering process for all categories. \citet{thu2021} performed clustering with an alignment strategy to generate pseudo labels for all unlabeled data to learn about known and novel categories. Furthermore, \citet{wscl} proposed a self-contrastive framework to discover fine-grained categories.
However, these methods cannot explicitly transfer category-specific knowledge from labeled data to unlabeled data, which can hurt model performance on known categories. Furthermore, they learn about known and novel categories in a coupled way, which can hurt model's generalization and discriminative ability. 

\subsection{Prototypical Learning}
Prototypical Learning (PL) assumes that each category can be represented by a single prototype in the feature space \citep{proto}. Most PL methods focused on matching instances to prototypes by computing distances between them. For example, \citet{proto2} performed cross-domain instance-to-prototype matching for Unsupervised Domain Adaptation (UDA). \citet{proto3} performed PL to jointly bridge the gap between domains and construct classifiers in the target domain for UDA. There were also some works that utilized PL for interpretability. For instance, \citet{interpretable1} performed interpretable image recognition by comparing image parts to learned prototypes. \citet{interpretable2} combined PL with decision trees to perform image recognition to address accuracy-interpretability trade-off. However, these methods simply assigned each instance into a single prototype based on one-hot pseudo labels, which can easily make models overfit on the noisy pseudo labels in the GCD setting.


\section{Method}
\subsection{Problem Statement}
Traditional classification models are developed based on a labeled dataset $\mathcal{D}^{l} = \{(x_{i},y_{i})|y_{i} \in \mathcal{Y}_{k}\}$, which contains only known categories $\mathcal{Y}_{k}$. However, in the real world, the deployed model may encounter unlabeled data $\mathcal{D}^{u} = \{x_{i}|y_{i} \in \{\mathcal{Y}_{k}, \mathcal{Y}_{n} \}\}$ which contains both known categories $\mathcal{Y}_{k}$ and novel categories $\mathcal{Y}_{n}$. So the aim of Generalized Category Discovery (GCD) is to recognize both known and novel categories based on $\mathcal{D}^{l}$ and $\mathcal{D}^{u}$. Finally, model performance will be tested on the testing set $\mathcal{D}^{t} = \{(x_{i},y_{i})|y_{i} \in \{\mathcal{Y}_{k}, \mathcal{Y}_{n} \}\}$.

\subsection{Approach Overview}
An overview of our \textit{Decoupled Prototypical Network} (DPN) is shown in Figure \ref{fig:picture1}, which contains five steps (marked in the figure).
First, we pretrain a feature extractor $F_{\theta}$ on both labeled and unlabeled data for representation learning. 
Second, we learn two sets of category prototypes based on $F_{\theta}$ for labeled and unlabeled data, respectively. 
Third, We match these two sets of prototypes to align known categories in labeled and unlabeled data, meanwhile decouple known and novel categories from unlabeled data with our proposed ``\textit{Alignment and Decoupling}" strategy.
Fourth, we propose \textit{Semantic-aware Prototypical Learning} (SPL) to learn more discriminative features for known and novel categories in a decoupled way. We also use the prototypes learned from the labeled data to transfer category-specific knowledge explicitly. 
Finally, we update category prototypes of labeled data with Exponential Moving Average (EMA) algorithm to capture the latest acquired knowledge. 

\subsection{Representation Learning}
We use the pretrained language model BERT \citep{bert} as our feature extractor $F_{\theta}: \mathcal{X} \rightarrow \mathbb{R}^{d}$. To learn semantic representations for current tasks, we fine-tune $F_{\theta}$ on both labeled and unlabeled data. We employ a joint pretraining loss following \citet{pretrain}, which contains Cross-Entropy (CE) loss for labeled data and Masked Language Modeling (MLM) loss \citep{bert} for all data:
\begin{equation}
     \mathcal{L}_{pre} = \mathcal{L}_{ce}(\mathcal{D}^{l}) + \mathcal{L}_{mlm}(\mathcal{D}^{l}, \mathcal{D}^{u})
\end{equation}
$\mathcal{D}^{l}$ and $\mathcal{D}^{u}$ are labeled and unlabeled dataset, respectively. Through pretraining, models can acquire various general knowledge for both known and novel categories and learn meaningful semantic representations for subsequent tasks.

\subsection{Learning Category Prototypes}
After pretraining, most existing works \citep{thu2020,thu2021,contras} generated pseudo labels for all unlabeled data to learn about known and novel categories in a coupled way. 
However, they ignored the different training goals for known and novel categories, and the coupled training manner made them difficult to transfer category-specific knowledge for known categories since they cannot align known categories in labeled and unlabeled data.
To mitigate above problems, we propose a novel ``\textit{Alignment and Decoupling}" strategy. Specifically, we learn two sets of category prototypes for labeled and unlabeled data, respectively. By aligning these two sets of prototypes, we can not only align known categories in labeled and unlabeled data to transfer category-specific knowledge explicitly in a category-to-category way, but also decouple known and novel categories from unlabeled data to acquire different knowledge about them with different training manners.

For labeled data, we take average of all instance embeddings belonging to the same category as labeled prototypes $P^{l} = \{\mu_{j}^{l}\}_{j=1}^{M}$, where $\mu_{j}^{l} = \frac{1}{|C_{j}|} \sum_{x_{i} \in C_{j}} F_{\theta}(x_{i})$ denotes the labeled prototype for category $j$, $C_{j}$ denotes a set of instances from category $j$ and $M=|\mathcal{Y}_{k}|$ is the number of known categories. For unlabeled data, we first perform KMeans clustering to get clusters $C^{u}=\{C^{u}_{1},C^{u}_{2},...,C^{u}_{K}\}$, where $K=|\mathcal{Y}_{k}|+|\mathcal{Y}_{n}|$ is the number of total categories. We presume prior knowledge of $K$ following previous works \citep{thu2021,contras} to make a fair comparison and we tackle the problem of estimating this parameter in the experiment. Then we take average of all instance embeddings belonging to the same cluster as unlabeled prototypes $P^{u} = \{\mu_{j}^{u}\}_{j=1}^{K}$, where $\mu_{j}^{u} = \frac{1}{|C_{j}^{u}|} \sum_{x_{i} \in C_{j}^{u}} F_{\theta}(x_{i})$.

\subsection{Alignment and Decoupling}
Instead of introducing extra data and parameters to train a binary classifier like previous two-stage methods \citep{twostage}, we propose to decouple known and novel categories in unlabeled data by aligning the learned prototypes $P^{u}$ and $P^{l}$. Since unlabeled data contain all known categories (following assumptions of previous works), we can find prototypes in $P^{u}$ which represent known categories. Intuitively, we think the closest prototypes in labeled and unlabeled data represent the same category. So we can formulate a bipartite matching problem to align known categories in the prototype sets $P^{u}$ and $P^{l}$.
Specifically, to find a bipartite matching between $P^{l}$ and $P^{u}$, we search for all possible permutation $\mathcal{P}_{all}$ for the set $P^{u}$ and get the optimal permutation $\mathcal{\hat{P}}$ by minimizing the total matching cost:
\begin{equation}
     \mathcal{\hat{P}} = \mathop{\arg\min}_{\mathcal{P}\in\mathcal{P}_{all}}\sum_{i=1}^{M} \mathcal{L}_{match}(\mu_{i}^{l}, \mu_{\mathcal{P}(i)}^{u})
\end{equation}
where $\mathcal{L}_{match}(\mu_{i}^{l}, \mu_{\mathcal{P}(i)}^{u})$ is point-to-point matching cost between the labeled prototype $\mu_{i}^{l}$ and the unlabeled prototype $\mu_{\mathcal{P}(i)}^{u}$. Here, we use the Euclidean distance as the matching cost to find the minimum distance matching:
\begin{equation}
\mathcal{L}_{match}(\mu_{i}^{l}, \mu_{\mathcal{P}(i)}^{u}) = \left\|\mu_{i}^{l}-\mu_{\mathcal{P}(i)}^{u} \right\|_{2}
\end{equation}

The bipartite matching problem can be solved efficiently with the Hungarian algorithm \citep{hungarian}. After that, we can get an optimal matching between $P^{l}$ and parts of $P^{u}$: $\{\mu_{j}^{l},\mu_{\mathcal{\hat{P}}(i)}^{u}\}_{i=1}^{M}$. So these matched prototypes in $P^{u}$ can be considered to represent known categories, denoting as $P^{uk} = \{\mu_{\mathcal{\hat{P}}(i)}^{u}\}_{i=1}^{M}$.
And the remaining unmatched prototypes in $P^{u}$ are seen to represent novel categories, denoted as $P^{un} = \{\mu_{\mathcal{\hat{P}}(i)}^{u}\}_{i=M+1}^{K}$.
Furthermore, the unlabeled data $\mathcal{D}^{u}$ can also be decoupled into two parts: $\mathcal{D}^{uk}$ for data with known categories and $\mathcal{D}^{un}$ for data with novel categories, which are decided by the clusters data belong to.

\subsection{Semantic-aware Prototypical Learning}
By decoupling known and novel categories, we can design specific training loss for them to achieve different training goals. Specifically, we propose \textit{Semantic-aware Prototypical Learning} (SPL) to learn about novel categories in an unsupervised manner. Then we transfer category-specific knowledge from labeled data to learn about known categories in a semi-supervised manner.
\subsubsection{Unsupervised Learning for Novel Categories.}
Since data with novel categories are unlabeled, we aim to explore some novel knowledge to improve model’s discriminative ability on these categories with unsupervised learning. Previous works focused on instance-level discrimination with pseudo-label training \citep{thu2021} or contrastive learning \citep{contras}. However, they ignored high-level semantics between instances and categories. To capture  high-level semantics, Prototypical Learning (PL) \citep{proto} was proposed to pull instances closer to prototypes they belong to and separate instances far away from other irrelevant prototypes:
\begin{equation}
\mathcal{L}_{pl} = -\frac{1}{n}\sum_{i=1}^{n}log\frac{e^{-d(F_{\theta}(x_{i}),\mu_{g}^{u})}}{\sum_{j=1}^{K}e^{-d(F_{\theta}(x_{i}),\mu_{j}^{u})}}
\end{equation}
where $x_{i}$ belongs to cluster $C^{u}_{g}$ and $\mu_{g}^{u}$ is the corresponding prototype. $d(\cdot,\cdot)$ is a distance function (e.g., Euclidean distance \citep{proto} or cosine distance \citep{pcl}).

However, PL simply divides each instance into a single prototype in a hard way and does not consider semantic relationships between instances and prototypes. Furthermore, since the mapping between instances and prototypes comes from pseudo-labels, PL is easily affected by pseudo-label noise. To mitigate above issues, we propose \textit{Semantic-aware Prototypical Learning} (SPL). Instead of assigning an instance into a single prototype based on the noisy pseudo labels, SPL assigns each instance into all prototypes using semantic similarity as weights:

\begin{equation}
\mathcal{L}_{spl} = \frac{1}{n}\sum_{i=1}^{n}\sum_{k=1}^{K^{\prime}}d(F_{\theta}(x_{i}),\mu_{k}^{u}) \frac{e^{s(F_{\theta}(x_{i}),\mu_{k}^{u})}}{\sum\limits_{j=1}^{K^{\prime}}e^{s(F_{\theta}(x_{i}),\mu_{k}^{u})}}
\end{equation}
where $n=|\mathcal{D}^{un}|$ is the number of unlabeled instances belonging to novel categories and $K^{\prime}$ is the number of novel categories, $d(\cdot,\cdot)$ is a distance function to pull instances and prototypes closer, $s(\cdot,\cdot)$ is a similarity function to evaluate semantic similarities between instances and prototypes. We use Euclidean distance as $d(\cdot,\cdot)$ and cosine similarity as $s(\cdot,\cdot)$ for novel categories in our paper. So Eq. (5) can be written as:
\begin{equation}
\mathcal{L}_{spl} = \frac{1}{n}\sum_{i=1}^{n}\sum_{k=1}^{K^{\prime}} \left \|F_{\theta}(x_{i})-\mu_{k}^{u} \right \|_{2}  \frac{e^{cos(F_{\theta}(x_{i}),\mu_{k}^{u})/\tau}}{\sum\limits_{j=1}^{K^{\prime}}e^{cos(F_{\theta}(x_{i}),\mu_{j}^{u})/\tau}}
\end{equation}
where $\tau$ is a temperature hyperparameter. By weighting with semantic similarities and assigning instances into prototypes in a soft way, SPL can capture meaningful semantic information between instances and prototypes and alleviate the effect of pseudo-label noise, especially for instances near decision boundaries. For simplicity, We denote the loss for novel categories as: 
\begin{equation}
\mathcal{L}_{novel} = \mathcal{L}_{spl}(\mathcal{D}^{un},P^{un})
\end{equation}

\begin{table}[t]
\centering
\begin{tabular}{lccccc}
\hline
Dataset & $|\mathcal{Y}_{k}|$ & $|\mathcal{Y}_{n}|$ & $|\mathcal{D}^{l}|$ & $|\mathcal{D}^{u}|$ & $|\mathcal{D}^{t}|$\\
\hline
BANKING         &  58    &  19     & 673    & 8,330    & 3,080\\
StackOverflow   &  15    &  5      & 1,350  & 16,650   & 1,000\\
CLINC           &  113   &  37     & 1,344  & 16,656   & 2,250\\
\hline

\end{tabular}
\caption{Statistics of datasets. $|\mathcal{Y}_{k}|$, $|\mathcal{Y}_{n}|$, $|\mathcal{D}^{l}|$, $|\mathcal{D}^{u}|$ and $|\mathcal{D}^{t}|$ represent the number of known categories, novel categories, labeled data, unlabeled data and testing data, respectively.}
\label{table1}
\end{table}

\begin{table*}
\centering

\begin{tabular}{lccccccccc}
\toprule
\multirow{2}*{Method} & \multicolumn{3}{c}{BANKING} &\multicolumn{3}{c}{StackOverflow} & \multicolumn{3}{c}{CLINC}\\ 
\cmidrule(r){2-4}  \cmidrule(r){5-7}  \cmidrule(r){8-10}
            &All    &Known    &Novel    &All    &Known    &Novel    &All    &Known    &Novel \\
\midrule
DeepCluster &13.95  &13.94  &13.99  &17.37  &18.22  &14.80  &26.92  &27.34  &25.67  \\
DCN         &17.85  &18.94  &14.35  &29.10  &28.94  &29.51  &29.64  &30.00  &28.45  \\
DEC         &19.30  &20.36  &15.84  &19.30  &20.36  &15.84  &19.99  &20.18  &19.40  \\
BERT        &21.29  &21.48  &20.70  &16.80  &16.67  &17.20  &34.52  &34.98  &33.16  \\
KM-GloVe    &29.18  &29.11  &29.39  &28.40  &28.60  &28.05  &51.64  &51.74  &51.50  \\
AG-GloVe    &30.09  &29.69  &31.29  &29.23  &28.49  &31.56  &44.70  &45.17  &43.20  \\
SAE         &38.05  &38.29  &37.27  &60.33  &57.36  &69.02  &46.59  &47.35  &44.24  \\

\midrule
Semi-DC   &50.73  &53.37   &42.63  &64.90  &66.13  &61.20  &74.52  &75.60  &71.34  \\
CDAC+    &53.09  &55.42  &46.01  &76.67  &77.51  &74.13  &69.75  &70.08  &68.77  \\
Self-Labeling    &56.19  &61.64  &39.56  &71.03  &78.53  &48.53  &72.69  &80.06  &49.65  \\
DTC      &56.56  &59.98  &46.10  &70.50  &80.93  &51.87  &76.42  &82.34  &58.95  \\
DAC       &63.63  &69.60  &45.44  &70.77  &76.13  &54.67  &84.42  &89.10  &70.59  \\
Semi-KM     &66.23  &73.62  &43.68  &73.13  &81.02  &49.47  &81.42  &89.03  &59.01  \\
LASKM    &67.55    &75.16  &44.34  &74.83  &82.00  &53.33  &79.26  &89.64  &48.66  \\
\midrule
\textbf{DPN} (\textbf{Ours})        &\textbf{72.96}  &\textbf{80.93}  &\textbf{48.60}  &\textbf{84.23}  &\textbf{85.29}  &\textbf{81.07}  &\textbf{89.06}  &\textbf{92.97}  &\textbf{77.54}  \\
Improvement    &+5.41    &+5.77  &+2.50  &+7.56  &+3.29  &+6.94  &+4.64  &+3.33  &+6.20  \\
\bottomrule

\end{tabular}
\caption{Model comparison results (\%) on testing sets. Average results over 3 runs are reported.}
\label{table2}
\end{table*}

\subsubsection{Semi-supervised Learning for Known Categories.}
For known categories, we aim to exploit knowledge acquired from labeled data to improve model’s generalization ability on these categories with semi-supervised learning. 
Previous approaches only used labeled data to transfer some general knowledge by pretraining models. However, they do not consider transferring high-level category knowledge explicitly for known categories to guide the pseudo-label training process, which can lose high-level semantic information and impair their model performance. To mitigate this issue, in addition to using SPL to learn some discriminative features, we further utilize the labeled prototypes $P^{l}$ to guide the pseudo-label training process and transfer category-specific knowledge explicitly as a regularization term:
\begin{equation}
\mathcal{L}_{reg} = \frac{1}{r}\sum_{i=1}^{r}\sum_{k=1}^{M} (1 - cos(F_{\theta}(x_{i}),\mu_{k}^{l}))  \frac{e^{cos(F_{\theta}(x_{i}),\mu_{k}^{l})/\tau}}{\sum\limits_{j=1}^{M}e^{cos(F_{\theta}(x_{i}),\mu_{j}^{l})/\tau}}
\end{equation}
where $r=|\mathcal{D}^{uk}|$ is the number of unlabeled instances belonging to known categories. We use cosine distance as the distance function here to acquire different knowledge from $P^{u}$.
By pulling instances closer to labeled prototypes using semantic similarity as weights, our model can transfer category-specific knowledge from labeled data to unlabeled data and further mitigate the risk of model overfitting on the noisy pseudo labels. To avoid catastrophic forgetting for knowledge acquired from labeled data, we also add cross-entropy loss during training. So the loss for known categories can be denoted as: 
\begin{equation}
\mathcal{L}_{known} = \mathcal{L}_{spl}(\mathcal{D}^{uk},P^{uk})  + \mathcal{L}_{ce}(\mathcal{D}^{l}) + \gamma \cdot \mathcal{L}_{reg}(\mathcal{D}^{uk},P^{l})
\end{equation}
where $\gamma$ is a weighting factor for the regularization term. In this way, we can fully utilize knowledge acquired from labeled data, labeled prototypes and unlabeled prototypes to improve model’s generalization ability on known categories in a semi-supervised manner.

\subsubsection{Total Loss.}
Overall, the training objective of our model can be formulated as:
\begin{equation}
\mathcal{L}_{dpn} = \mathcal{L}_{novel} + \mathcal{L}_{known}
\end{equation}

\subsection{Updating Category Prototypes}
Despite of utilizing labeled prototypes to guide the decoupling and pseudo-label training process, we also update them periodically to capture the latest acquired knowledge so that they can generalize better to unseen instances. However, directly overwriting the original values can lead to less consistent representations for prototypes and make them easily affected by outliers, which will make training unstable. So we use Exponential Moving Average (EMA) algorithm to learn more robust prototypes for labeled data:
\begin{equation}
\mathcal{P}_{t+1}^{l} \leftarrow \alpha \cdot \mathcal{P}_{t}^{l} + (1 - \alpha) \cdot \mathcal{P}_{t+1}^{l}
\end{equation}
where $\alpha$ is a momentum factor. By adjusting $\alpha$, we can learn more consistent and robust prototype representations. Here, we fix unlabeled prototypes ${P}^{u}$ to avoid the time-consuming clustering process and the influence of randomly permuted cluster indices, without influencing our model performance.
After training, we apply a clustering algorithm (e.g., KMeans) to obtain cluster assignments for testing data.

\section{Experiments}
\subsection{Experimental Setup}
\subsubsection{Datasets.}
We evaluate our model on three benchmark datasets.
\textbf{BANKING} is an intent classification dataset in the banking domain released by \citet{banking}.
\textbf{StackOverflow} is a technical question classification dataset processed by \citet{stack}.  
\textbf{CLINC} is a text classification dataset with diverse domains released by \citet{clinc}. Statistics of these datasets can be found in Table \ref{table1}.

\subsubsection{Comparison Methods.} 
We compare our model with various baselines and state-of-the-art (SOTA) methods.

\noindent \textbf{Unsupervised Methods.}\quad  (1) KM-GloVe: KMeans \citep{km} with GloVe embeddings \citep{glove}. (2) AG-GloVe: Agglomerative Clustering \citep{ag} with GloVe embeddings. (3) SAE: Stacked Auto Encoder. (4) DEC: Deep Embedding Clustering \citep{dec}. (5) DCN: Deep Clustering Network \citep{dcn}. (6) KM-BERT: KMeans with BERT embeddings \citep{bert}. (7) DeepCluster: Deep Clustering \citep{deepcluster}.

\noindent \textbf{Semi-supervised Methods.}\quad (1) KM-Semi: KMeans with BERT pretrained on labeled data. (2) DeepCluster-Semi: Deep Clustering pretrained on labeled data. (3) DTC: Deep Transfer Clustering \citep{dtc}. (4) CDAC+: Constrained Adaptive Clustering \citep{thu2020}. (5) DAC: Deep Aligned Clustering \citep{thu2021}. (6) Self-Labeling: Self-Labeling Framework \citep{selflabel}. (7) LASKM: Label Assignment with Semi-supervised KMeans \citep{gcd}.

\begin{table}
\centering
\begin{tabular}{lccc}
\toprule
\textbf{Model} & All & Known & Novel \\
\midrule  
Ours & 84.23 & 85.29 & 81.07\\
\midrule
w/o Cross Entropy      & 83.83 & 85.02 & 80.26 \\
w/o EMA                & 82.50 & 83.87 & 78.40 \\
w/o Decoupling         & 78.77 & 78.53 & 79.47 \\
w/o Soft Assignment    & 75.10 & 75.33 & 74.40 \\
w/o Semantic Weights   & 35.70 & 33.73 & 41.60 \\
\bottomrule

\end{tabular}
\caption{Results (\%) of different model variants.}
\label{table3}
\end{table}

\subsubsection{Evaluation Metrics.}
We measure accuracy between ground-truth labels and model's cluster assignments on the testing set with Hungarian algorithm \citep{hungarian}.
(1) All: accuracy for all instances.
(2) Known: accuracy for instances with known categories.
(3) Novel: accuracy for instances with novel categories.

\subsubsection{Implementation Details.}
We use the pretrained BERT model (bert-base-uncased) implemented by Pytorch \citep{huggingface} and adopt its suggested hyper-parameters. We pretrain model on all Transformer layers and only fine-tune on the last three Transformer layers to speed up calculation. We use the \textit{[CLS]} token of the last Transformer layer as instance feature. For model optimization, we use AdamW optimizer. 
Early stopping is used during pretraining, which is decided by model performance on the validation set which only contains known categories. 
For comparison methods, we use the implementations and hyper-parameters in their original papers. And some implementations are based on \citet{textoir}.
For hyper-parameters, $\gamma$ is set to \{10, 10, 90\} for BANKING, CLINC and StackOverflow dataset, respectively. $\alpha$ is set to 0.9 and $\tau$ is set to 0.07. 
Training epochs for StackOverflow, BANKING and CLINC dataset are set to \{10, 60, 80\}.
The wait patience for early stopping is set to 20. The learning rate for pretraining is set to $5e^{-5}$ and the learning rate for training is set to $1e^{-5}$.
For masked language modeling, the mask probability is set to 0.15 following previous works.

\subsection{Experimental Results}
\subsubsection{Main Results.} 
The results are shown in Table \ref{table2}. Overall, our model outperforms all comparison methods on all datasets and evaluation metrics with a large margin. 
First, our model achieves the best results on accuracy for all instances (denoted as `All'). Specifically, our model outperforms the SOTA model by \textbf{5.87\%} of average accuracy on three benchmark datasets, which can reflect the effectiveness of our model on both known and novel categories. 
Second, our model achieves the best performance on accuracy for known categories (denoted as `Known').
In detail, our model improves average accuracy of known categories by \textbf{4.13\%} over the SOTA method since we decouple known categories from unlabeled data and fully utilize acquired knowledge to learn about them in a semi-supervised manner. By aligning prototypes learned from labeled and unlabeled data, our model can decouple known and novel categories from unlabeled data effectively, so that we can obtain different knowledge with different training approaches. Furthermore, by aligning known categories in labeled and unlabeled data, our model can transfer category-specific knowledge from labeled data to unlabeled data explicitly through prototypes. 
Third, our model also gets the best performance on accuracy for novel categories (denoted as `Novel'). More specifically, average improvement of \textbf{5.21\%} can be seen on the accuracy of novel categories compared with the SOTA method, due to our \textit{Semantic-aware Prototypical Learning} (SPL). By measuring semantic similarities between instances and prototypes, our model can assign each instance into different prototypes in a soft manner, which can capture semantic information and mitigate the effect of pseudo-label noise.

\begin{figure}
\centering
\includegraphics[width=8cm, height=4.7cm]{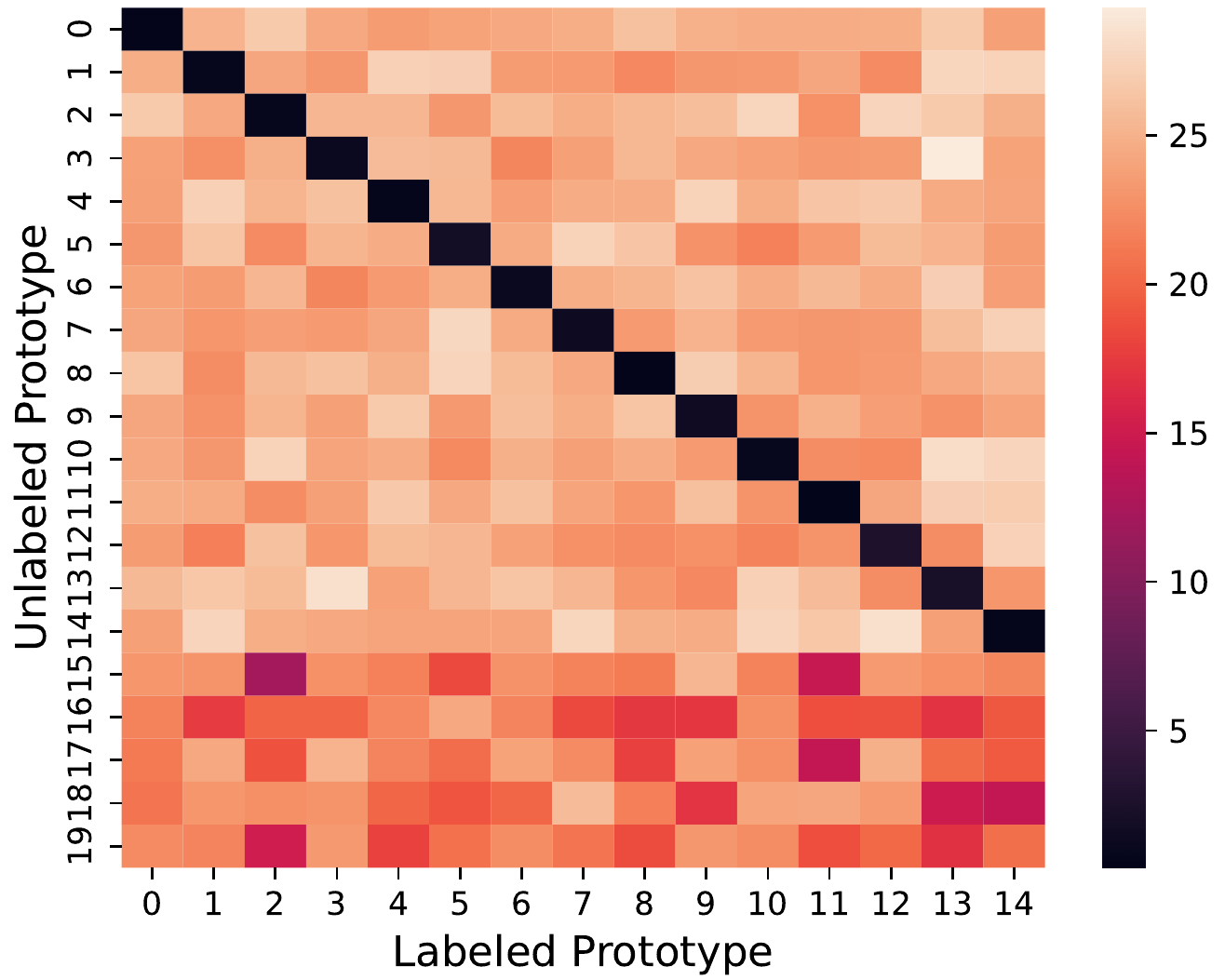}
\caption{Distances between 15 labeled prototypes and 20 aligned unlabeled prototypes. Darker colors represent closer distances.} 
\label{fig:picture3}
\end{figure}

\begin{figure*}
\centering
\includegraphics[width=15.5cm, height=5cm]{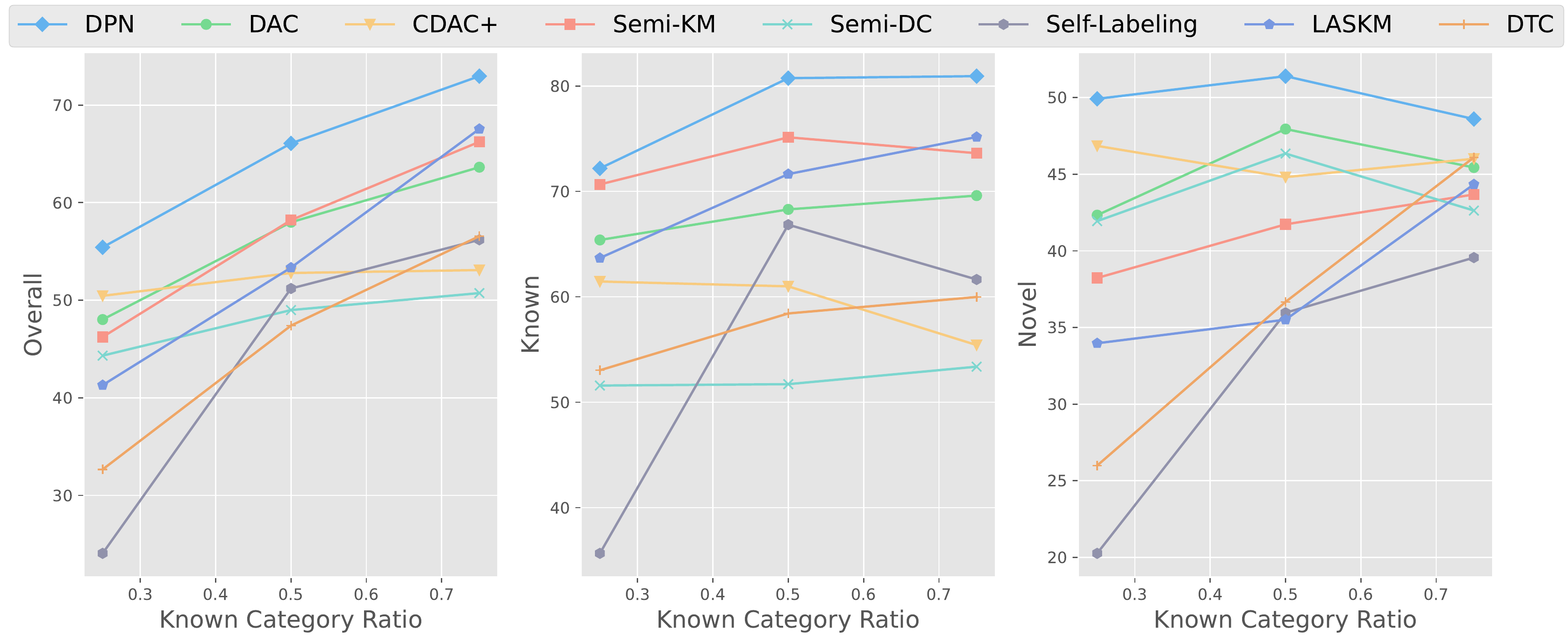}
\caption{Effect of known category ratio on the BANKING dataset.} 
\label{fig:picture4}
\end{figure*}

\begin{figure}
\centering
\subfigure[Semi-KM]{
\begin{minipage}[t]{0.5\linewidth}
\centering
\includegraphics[width=1.3in,height=2.25cm]{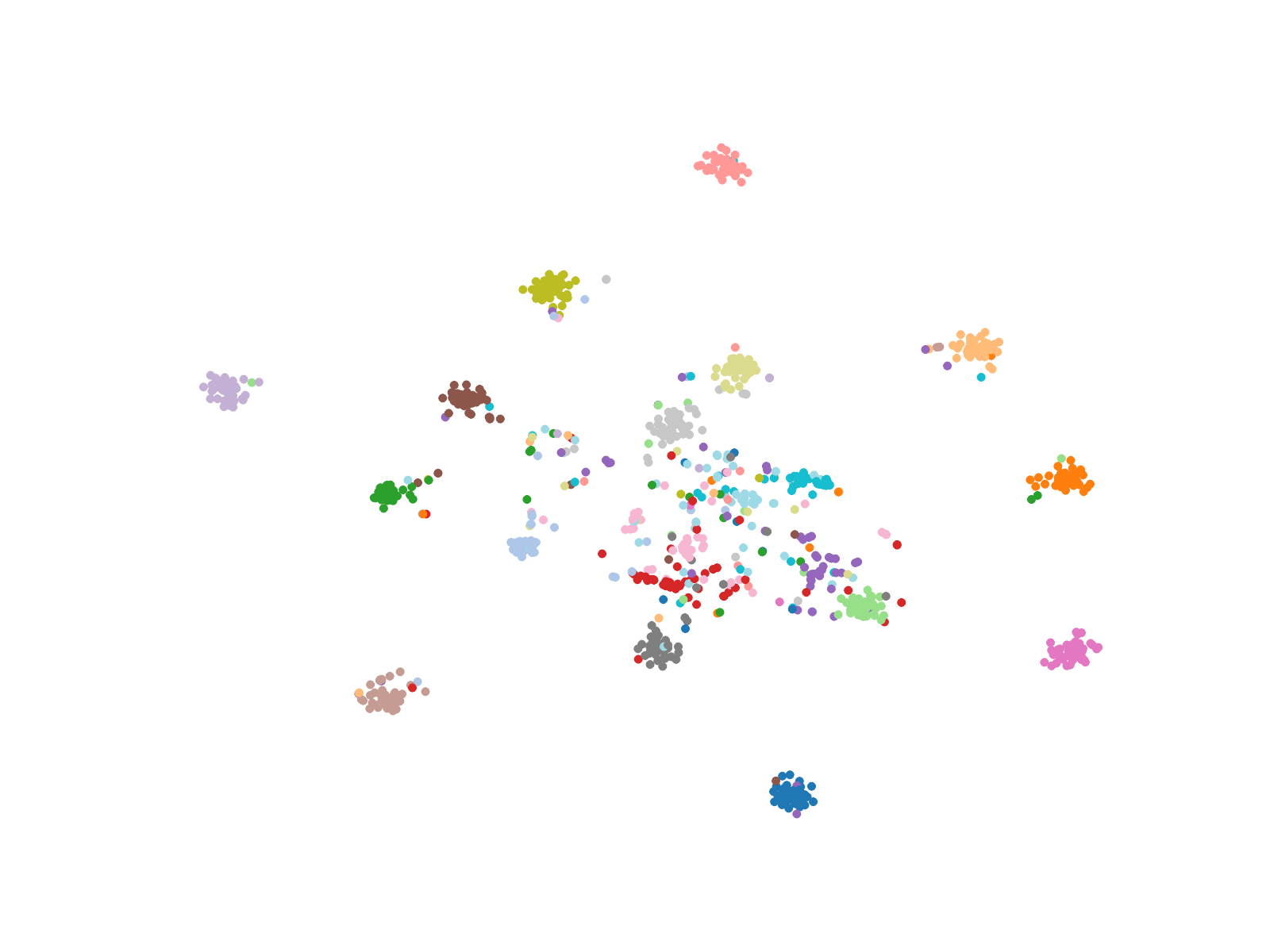}
\end{minipage}%
} \hspace{-10mm}
\subfigure[CDAC+]{
\begin{minipage}[t]{0.5\linewidth}
\centering
\includegraphics[width=1.3in,height=2.25cm]{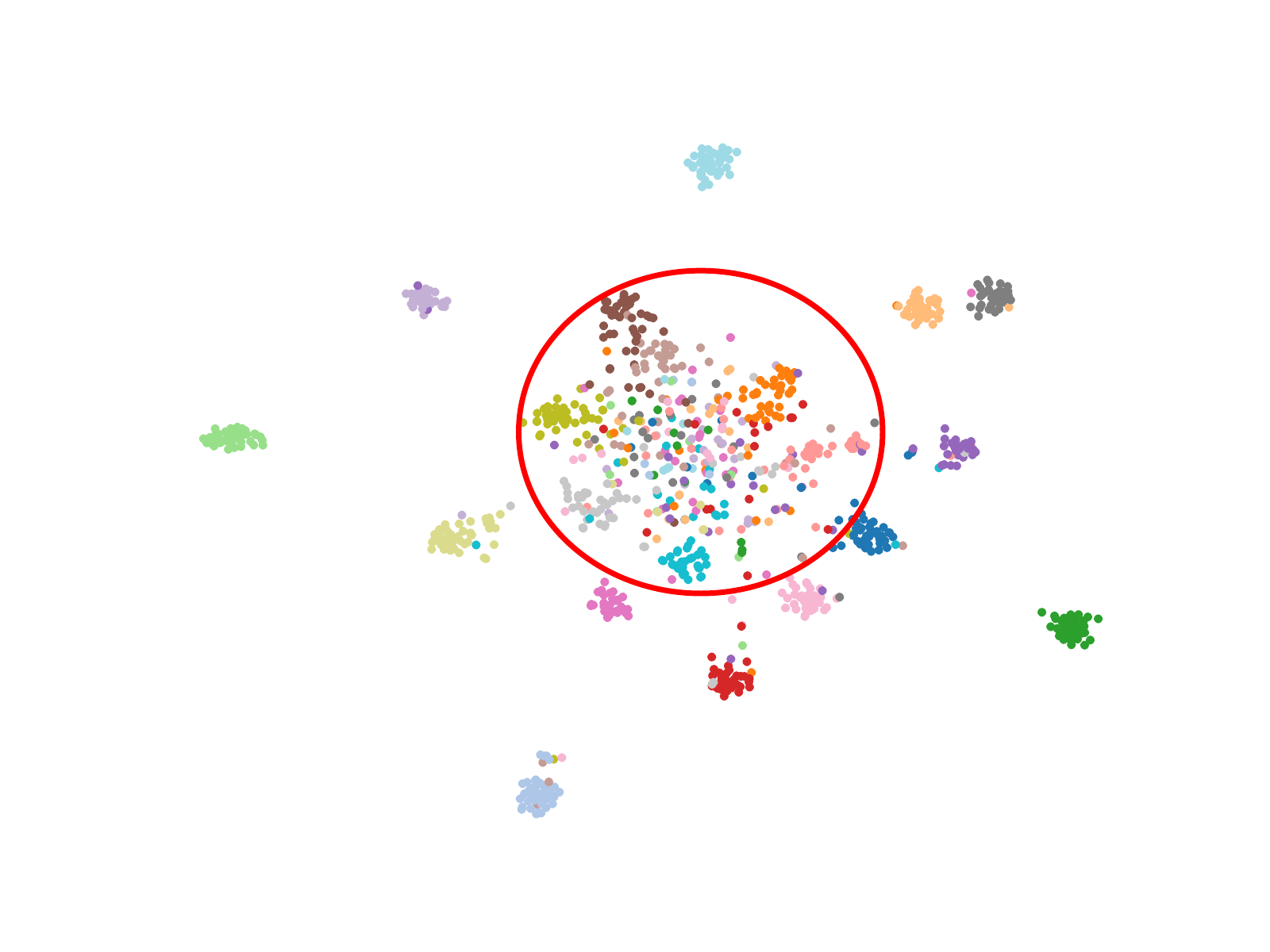}
\end{minipage}%
}%

\subfigure[LASKM]{
\begin{minipage}[t]{0.5\linewidth}
\centering
\includegraphics[width=1.3in, height=2.25cm]{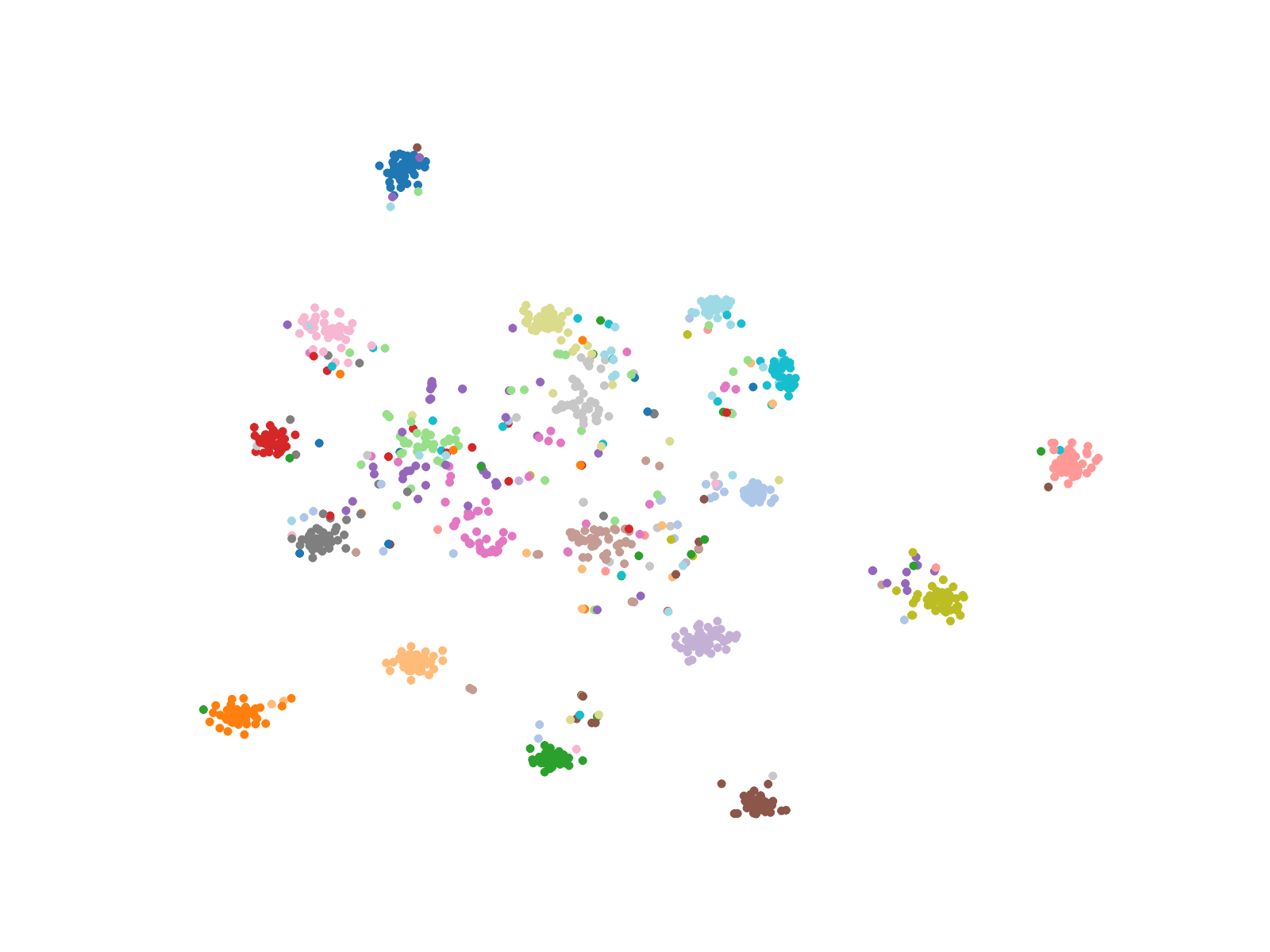}
\end{minipage}
} \hspace{-10mm}
\subfigure[DPN]{
\begin{minipage}[t]{0.5\linewidth}
\centering
\includegraphics[width=1.3in, height=2.25cm]{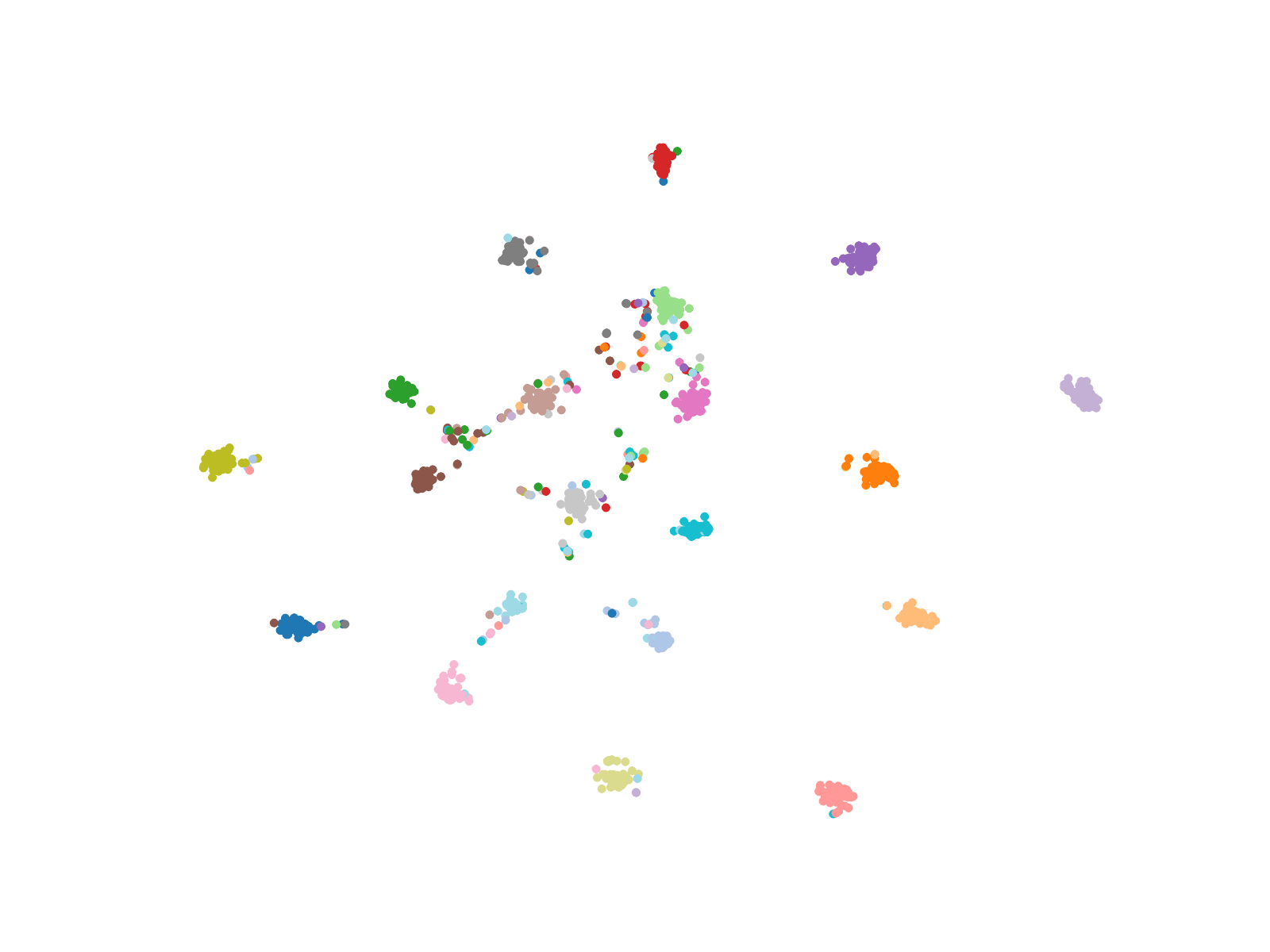}
\end{minipage}
}%
\centering
\caption{The t-SNE visualizations of embeddings.}
\label{fig:picture5}
\end{figure}

\subsubsection{Ablation Study.}
The performance of variants of our model on the StackOverflow dataset is shown in Table \ref{table3}. Overall, removing different components from our model will affect model performance more or less, which can show the effectiveness of different components of our model. Specifically, (1) Removing \textbf{Cross Entropy} loss $\mathcal{L}_{ce}$ in Eq. (9) has minimal influence since our model can also obtain knowledge for known categories through labeled prototypes. (2) Removing \textbf{EMA} updating for labeled prototypes in Eq. (11) will lead to inconsistent representations for prototypes and hurt model performance. (3) Removing \textbf{Decoupling} will prevent the category-specific knowledge transfer in Eq. (8) and hurt model performance on both known and novel categories. (4) Removing \textbf{Soft Assignment} in Eq. (5) can easily make models overfit on the noisy pseudo labels generated by clustering and damage model performance. (5) Removing \textbf{Semantic Weights} $s(F_{\theta}(x_{i}),\mu_{k}^{u})$ in Eq. (5) can lose semantic information between instances and prototypes and largely degrade model performance.

\subsubsection{Effectiveness of Alignment and Decoupling Strategy.}
To investigate the effectiveness of our \textit{Alignment and Decoupling} strategy, we visualize the heat map of distances between labeled prototypes and aligned unlabeled prototypes on the StackOverflow dataset in Figure \ref{fig:picture3}.
We can see that our strategy can align these two sets of prototypes effectively. 
Since prototypes correspond to categories, our model can align known categories in labeled and unlabeled data in a category-to-category way (the black grids) to transfer category-specific knowledge. Meanwhile, our model can decouple known and novel categories from unlabeled data, so that we can obtain different knowledge for them.

\subsubsection{Estimating the Number of Categories.}
Here, we use the algorithm proposed in DAC \citep{thu2021} to tackle the problem of estimating the number of categories \textit{K} from unlabelled dataset $\mathcal{D}^{u}$. The results are reported in Table \ref{table8}. We can see that our model achieves lower error rates on all datasets, which means that our model can learn better representations to estimate the number of categories.

\begin{table}
\centering
\begin{tabular}{lccc}
\toprule
 & CLINC & BANKING  & StackOverflow \\
\midrule
Ground Truth & 150      & 77    & 20 \\
\midrule
DAC          & 130      & 66    & 15 \\
Ours         & 137      & 67    & 18 \\
Error        & 8.7\%    & 13.0\%    & 10.0\% \\
\bottomrule

\end{tabular}
\caption{Estimation of the number of categories.}
\label{table8}
\end{table}

\subsubsection{Influence of Known Category Ratio.}
To investigate the influence of known category ratio on model performance, we vary 
it in the set \{0.25, 0.50, 0.75\}. As shown in Figure \ref{fig:picture4}, our model achieves the best performance under different settings on all evaluation metrics, which can show effectiveness and robustness of our model.

\subsubsection{Feature Visualization.}
In Figure \ref{fig:picture5}, we use t-SNE to illustrate embeddings learned by different methods on the StackOverflow dataset. 
It clearly shows that our model can learn more separable features for different categories than comparison methods, which can further demonstrate the effectiveness of our model.

\section{Conclusion}
In this paper, we propose \textit{Decoupled Prototypical Network} for Generalized Category Discovery. 
Inspired by the different training goals for known and novel categories, we propose to decouple them from unlabeled data to acquire different knowledge. 
Then we formulate a bipartite matching problem for prototypes to align known categories in labeled and unlabeled data to transfer category-specific knowledge, meanwhile decouple known and novel categories from unlabeled data to acquire different knowledge.
Furthermore, we propose \textit{Semantic-aware Prototypical Learning} (SPL) to learn more discriminative features from unlabeled data. By assigning each instance into different prototypes using semantic similarities as weights, our model can capture meaningful semantic information to learn about known and novel categories and alleviate the effect of noisy pseudo labels.
Experimental results on three benchmark datasets show that our model outperforms state-of-the-art methods by a large margin.

\section{Acknowledgments}
This work was supported by National Key Research and Development Program of China (2020AAA0108800), National Natural Science Foundation of China (62137002, 61721002, 61937001, 61877048, 62177038, 62277042). Innovation Research Team of Ministry of Education (IRT\_17R86), Project of China Knowledge Centre for Engineering Science and Technology. MoE-CMCC ``Artifical Intelligence''  Project (MCM20190701), Project of Chinese academy of engineering ``The Online and Offline Mixed Educational ServiceSystem for 'The Belt and Road’ Training in MOOC China''. ``LENOVO-XJTU'' Intelligent Industry Joint Laboratory Project.

\bibliography{reference}
\end{document}